\documentclass[11pt]{article}

\usepackage{acl}

\usepackage{times}
\usepackage{latexsym}
\usepackage[T1]{fontenc}
\usepackage[utf8]{inputenc}
\usepackage{microtype}
\usepackage{inconsolata}
\usepackage{booktabs}
\usepackage{float}
\usepackage{multirow}
\usepackage{amsmath}
\usepackage{graphicx}
\usepackage{xcolor}

\usepackage[capitalize,noabbrev,nameinlink]{cleveref}

\crefname{appendix}{Appendix}{Appendices}
\Crefname{appendix}{Appendix}{Appendices}

\definecolor{mylinkblue}{RGB}{25,25,112}
\hypersetup{
    pdftitle={Positional Failures in Long-Context LLMs: A Blind Spot in Reasoning Benchmarks},
    pdfauthor={Chuyifei Zhang, Hongyu Cui, Xiaowen Huang, Jitao Sang},
    pdfsubject={Position-controlled evaluation for long-context reasoning benchmarks},
    pdfkeywords={long-context language models, evaluation benchmarks, positional reasoning, lost-in-the-middle, vendor model card audit},
    colorlinks=true,
    linkcolor=mylinkblue,
    citecolor=mylinkblue,
    urlcolor=mylinkblue
}

\title{Positional Failures in Long-Context LLMs:\\
A Blind Spot in Reasoning Benchmarks}

\author{
  Chuyifei Zhang \\
  Beijing Jiaotong University \\
  \texttt{24222058@bjtu.edu.cn}
  \And
  Hongyu Cui \\
  Central South University \\
  of Forestry and Technology \\
  \texttt{zoean@csuft.edu.cn}
  \AND
  Xiaowen Huang\thanks{Corresponding author.} \\
  Beijing Jiaotong University \\
  \texttt{xwhuang@bjtu.edu.cn}
  \And
  Jitao Sang \\
  Beijing Jiaotong University \\
  \texttt{jtsang@bjtu.edu.cn}
}

\begin{document}
\maketitle

\begin{abstract}
Position-controlled evaluation is widely adopted for retrieval tasks
such as Needle-in-a-Haystack and RULER. Mainstream reasoning
benchmarks, by contrast, fail to control the positional placement of
target tasks within long contexts. We audit 11 popular long-context
benchmarks and find that none jointly controls task position, filler
content, and context length for reasoning. An audit of four recent
flagship long-context releases finds no main result-table entry
for NIAH, RULER, or LongBench-family benchmarks, while agentic
and coding benchmarks appear in main result-tables across all four.
These audited vendor result tables therefore provide little direct
visibility into position-controlled reasoning behavior. In response
to these gaps in benchmark design and vendor evaluation practice,
we propose Context Rot Evaluation (CRE), a controlled framework
that systematically varies all three factors. We evaluate nine LLMs
on GSM8K and ARC-Challenge across two rounds: an initial five-model
set and four newer vendor releases. Models can drop sharply when
the target task moves from the end of the context to the middle,
and the drop grows worse with context length for vulnerable models.
MiMo-v2-Flash, for example, drops 88pp at 64K under \textit{with\_solutions}
filler, with middle accuracy falling to 8\%. The newer releases show
smaller middle-position drops under the same filler. At 64K, three
of the four stay within $\pm$6pp of end-position accuracy. MiMo-V2.5-Pro
narrows the drop from MiMo-v2-Flash's 88pp to 32pp. Under
\textit{questions\_only\_v2} filler, middle-position drops persist across all
four (range $-16$pp to $-56$pp across 8K, 32K, and 64K). At 8K, a diagnostic probe that adds a
target-task copy at the end brings middle-position accuracy within
$\pm$4pp of end-position baseline across all nine models. This pattern
is consistent with a positional explanation. In the initial
five-model set, 76\% of middle-position errors match surrounding filler
text versus 22\% at the end position, consistent with filler-answer
interference as a dominant error mode. These
results expose a structural evaluation gap in current reasoning
benchmark design and vendor evaluation practice: when task position
is not controlled, positional vulnerabilities that grow with context
length cannot be measured as a separate effect.
\end{abstract}

\section{Introduction}
\label{sec:intro}

Modern long-context LLMs claim effective context windows of 128K tokens or
more. Whether these claims hold up in practice depends on how we evaluate
them. For retrieval tasks, position control is already standard:
Needle-in-a-Haystack \citep{kamradt2023needle} sweeps answer position
across context depth, and RULER \citep{hsieh2024ruler} adds configurable
task complexity. By contrast, mainstream long-context reasoning benchmarks follow a
different practice: task position is rarely controlled or explicitly
reported. Representative suites including SCROLLS \citep{shaham2022scrolls}, LongBench
\citep{bai2024longbench}, L-Eval \citep{an2023leval}, and LooGLE
\citep{li2023loogle} evaluate models on natural documents whose
task-relevant information sits at positions inherited from the source
rather than chosen experimentally.

This asymmetry between retrieval and reasoning benchmarks introduces
an evaluation gap. Natural-document reasoning benchmarks can only
report aggregated model performance on long-context tasks, but they
cannot disentangle success or failure from task position. If a model fails at some positions but not others,
those failures stay uncharacterized when benchmarks do not control task
position. None of the surveyed benchmarks jointly controls task position,
filler content, and context length for reasoning tasks. Whether this gap
matters empirically remains an open question.

We then audit how four recent flagship releases (DeepSeek-V4-Pro,
MiMo-V2.5-Pro, Kimi-K2.6, and GLM-5.1) report their own evaluation
results, and the same gap appears in vendors' own tables.
Across the four releases, the seven long-context benchmark
families tracked in this audit (NIAH, RULER, LongBench, HELMET,
$\infty$Bench, BABILong, LOFT) are absent from all twenty-eight
main result-table audit cells. By contrast, each vendor's headline
comparison includes SWE-Bench and browser- or terminal-oriented
agentic/coding benchmarks, such as BrowseComp, Terminal-Bench,
ClawEval, or GDPVal. DeepSeek-V4-Pro, for example, reports
SWE-Bench Verified 80.6, BrowseComp 83.4, and Terminal-Bench 67.9
in its main Instruct table. NIAH and RULER are absent, and
LongBench-V2 (51.5 EM) appears only in the Base eval table.
GLM-5.1 reports NIAH and RULER inside a 9B-proxy architecture
ablation for sparse attention, not as primary positional fidelity
claims for the deployed product. The full vendor disclosure matrix
is in \cref{app:vendor-audit}.

These four vendors lead their main result-tables with agent and coding
task scores rather than with positional reasoning scores. Strong
scores on agent and coding benchmarks would not, on their own,
characterize position-controlled reasoning behavior.
Position-controlled benchmarks isolate the variable that these
vendor tables leave uncharacterized: how reasoning performance
changes with task position. This measurement gap motivates CRE.

Our work bridges these gaps. We first audit 11 long-context benchmarks
(\cref{sec:related}) and document a structural asymmetry:
position-controlled evaluation is well established for retrieval but
largely absent from mainstream reasoning benchmark design. Building on
this audit, we propose Context Rot Evaluation (CRE), a controlled
framework that systematically varies three factors---task position,
filler content, and context length---for reasoning tasks. CRE evaluates nine LLMs across two rounds: an initial set
(Qwen 2.5-7B-Instruct, MiMo-v2-Flash, GLM-4.7-FlashX, DeepSeek-V3.2
in reasoning mode, and Kimi k2.5) plus four newer vendor releases
(DeepSeek-V4-Pro, MiMo-V2.5-Pro, Kimi-K2.6, and GLM-5.1), tested across three
context tiers (8K, 32K, 64K) and three filler types (worked solutions,
unanswered questions, neutral Wikipedia/news prose). The reasoning tasks
are GSM8K \citep{cobbe2021gsm8k} (math word problems) and ARC-Challenge
\citep{clark2018arc} (science multiple choice). With this setup, we
characterize a model-dependent vulnerability spectrum and analyze
diagnostic probe behavior across filler types and context lengths.

End-to-middle drops reach 94pp on GSM8K under structured-filler conditions;
76\% of middle-position errors match filler content versus 22\% at the end,
consistent with filler-answer interference as a dominant error mode.

Our contributions are:
\begin{itemize}
    \itemsep0em
    \item A systematic position audit of 11 long-context benchmarks:
    position-aware evaluation is comparatively mature for retrieval
    but largely unaudited for mainstream reasoning; synthetic
    exceptions like BABILong and LongReason do not close this gap
    (\cref{sec:related,app:benchmark-audit}).
    \item Evidence that task position, filler content, and context
    length jointly shape reasoning failures. End-to-middle drops
    reach 94pp on GSM8K, with directional cross-domain support
    on ARC-Challenge (range $-6$pp to $-40$pp); benchmarks that
    do not control task position miss these effects
    (\cref{sec:results}).
    \item A model-dependent vulnerability spectrum: some models drop
    sharply at the middle position, others stay nearly immune.
    Diagnostic probe behavior is consistent with a positional explanation
    of these failures (\cref{sec:results-probes}).
\end{itemize}

\section{Related Work}
\label{sec:related}

Position effects in long-context LLMs have been documented since
\citet{liu2023lostinmiddle}. They showed that multi-document QA models attend
less to information in the middle of the context than to information at the
beginning or end. This result motivated a line of synthetic retrieval-style
benchmarks that manipulate position directly.

Needle-in-a-Haystack \citep{kamradt2023needle} established the basic protocol
of sweeping answer position across context depth. RULER
\citep{hsieh2024ruler} extended this idea into a configurable synthetic suite
covering retrieval, variable tracking, and aggregation tasks. Later benchmarks
such as LongPiBench \citep{tian2025longpibench}, NoLiMa
\citep{modarressi2025nolima}, and Counting-Stars
\citep{song2025countingstars} further expanded position-aware evaluation with
multi-evidence settings, uniform position grids, and depth-alignment analyses.
Counting-Stars includes a reasoning-labeled subtask, but it remains a
deliberately simplified synthetic benchmark centered on multi-evidence
collection. The authors themselves note that the task does not require
summation. Taken together, these studies show that positional effects are
measurable when position is treated as a first-class variable. But they remain
primarily controlled synthetic benchmarks for retrieval-like or proxy tasks,
not audits of mainstream long-context reasoning evaluation.

By contrast, the benchmarks that dominate long-context reasoning and NLU
evaluation usually do not control or report the position of target
information. SCROLLS \citep{shaham2022scrolls}, ZeroSCROLLS
\citep{shaham2023zeroscrolls}, L-Eval \citep{an2023leval}, LongBench
\citep{bai2024longbench}, LooGLE \citep{li2023loogle}, and 100-LongBench
\citep{yang2025longbench100} evaluate models on realistic long-document tasks.
The location of task-relevant evidence comes from natural documents rather than
experimental control, so these benchmarks cannot isolate whether success or
failure depends on where the task appears.
InfiniteBench \citep{zhang2024infinitebench} shows this split within one
benchmark: it combines position-controlled retrieval subtasks (PassKey
variants) with natural-document reasoning subtasks whose positions are not
controlled.

The closest precedent to our audit is \citet{buchmann2024attribute}. They
study six long-document attribution datasets and report that response quality
tends to decrease when evidence appears later in the document. Its scope is
narrower than ours: it focuses on attribution benchmarks and does not examine
whether mainstream reasoning benchmark construction systematically leaves
positional vulnerabilities unmeasured.

Pos2Distill \citep{wang2025pos2distill} is also closely related. It explicitly
distinguishes retrieval and reasoning manifestations of position bias, but it
does so in the service of mitigation rather than benchmark audit. LongReason
\citep{ling2025longreason} provides a synthetic reasoning-side precedent for
position-aware evaluation through a binary ablation on final-inquiry
placement. It shows that reasoning performance can depend on whether the final
inquiry appears before or after the background context. However, its
positional control remains coarse-grained and does not amount to a
cross-benchmark audit or a joint position-content-length evaluation. Together,
these works strengthen the case that positional effects matter for reasoning,
but they do not eliminate the benchmark-design gap targeted by CRE.

BABILong \citep{kuratov2024babilong} is the closest earlier synthetic
reasoning precedent: it evaluates reasoning over supporting facts in long
distractor text but treats position as supplementary rather than as its
primary evaluation axis. RULER's Variable Tracking task approaches
coreference-style reasoning but evaluates information localization rather
than the multi-step natural-language reasoning targeted by SCROLLS or
LongBench.

Against this backdrop, CRE contributes a missing evaluation layer. Rather than
introducing another position-aware synthetic task, we combine a
cross-benchmark position audit (\cref{app:benchmark-audit}) with a controlled
evaluation framework for reasoning tasks. By jointly manipulating task
position, filler content, and context length on GSM8K and ARC-Challenge across
nine models, CRE exposes an evaluation gap left uncharacterized by
retrieval-centered position benchmarks and natural-document reasoning
benchmarks alike. The phrase ``context rot'' has also appeared in
industry write-ups (e.g., Chroma Research) characterizing
retrieval-side context degradation; CRE addresses a distinct,
reasoning-side question.

\section{Method}
\label{sec:method}

\subsection{CRE Framework}
\label{sec:method-framework}

\begin{figure*}[t]
\centering
\includegraphics[width=0.9\textwidth]{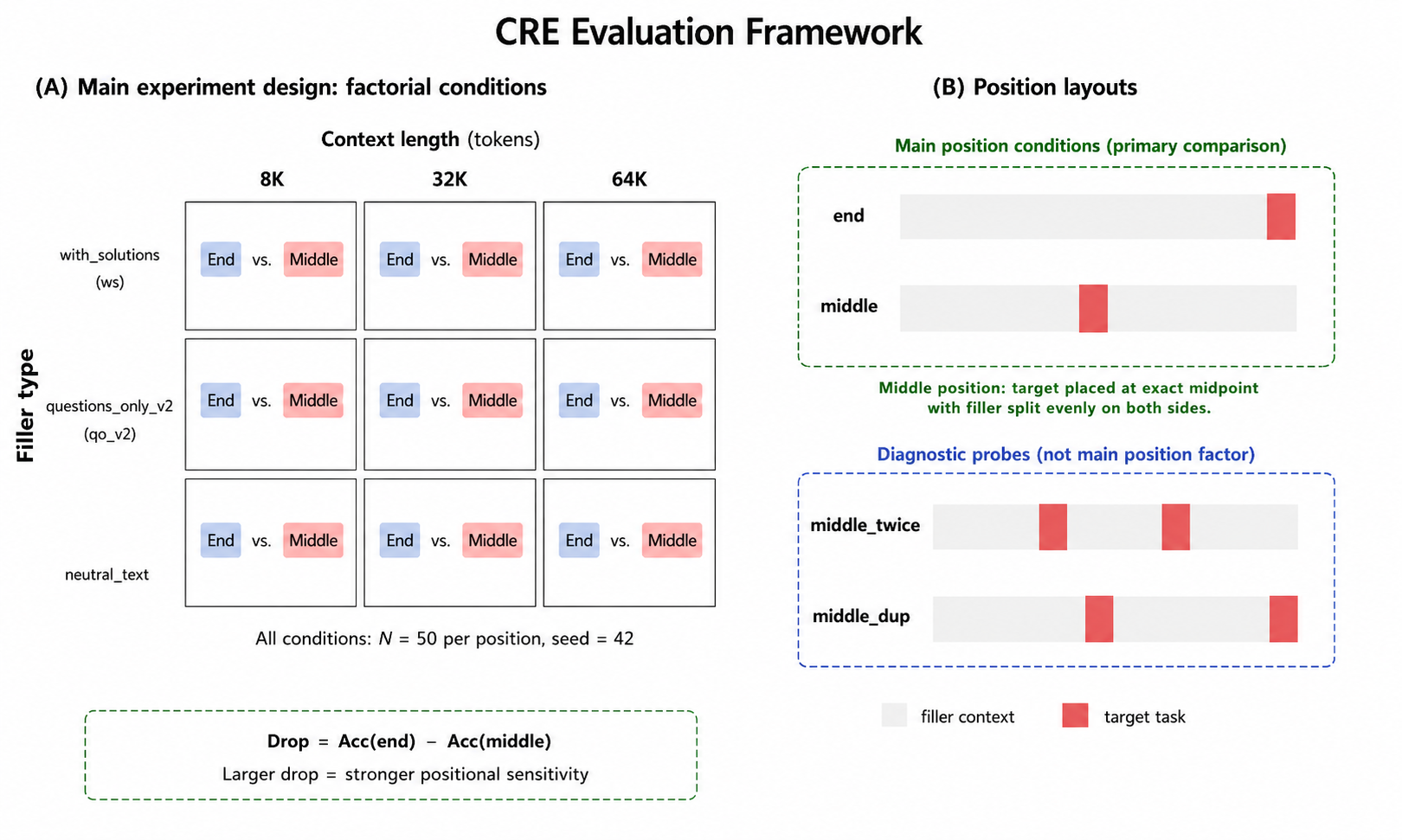}
\caption{CRE evaluation framework. (A) Main experiment design: a $3 \times 3$ factorial of filler types (\textit{with\_solutions}, \textit{questions\_only\_v2}, \textit{neutral\_text}) and context lengths (8K, 32K, 64K); each cell compares end vs.\ middle position with $N=50$ per condition (seed=42). (B) Position layouts: end and middle are the main comparison; middle\_twice and middle\_dup serve as diagnostic probes evaluated at 8K only.}
\label{fig:framework}
\end{figure*}

CRE (Context Rot Evaluation) is a controlled evaluation framework that treats task position, filler content, and context length as three independent variables (\cref{fig:framework}). For each reasoning task instance, we construct a prompt by embedding the target question among filler content at a specified position within a specified token budget. A full factorial design varies each axis independently, holding the other two fixed, so accuracy differences can be compared while holding the other axes fixed.

We evaluate four position variants: \textit{end} (target task at the end of the context, immediately preceding the model's response), \textit{middle} (target task in the middle of the context, with filler on both sides), \textit{middle\_twice} (target task repeated twice in the middle, with no copy at the end), and \textit{middle\_dup} (target task duplicated once in the middle and once at the end). The end and middle conditions isolate the primary positional effect; middle\_twice and middle\_dup are diagnostic probes used in \cref{sec:results-probes}. We define \textit{drop} as middle-position accuracy minus end-position accuracy at the same tier and filler type, so a more negative drop indicates greater positional sensitivity.

\subsection{Filler Content}
\label{sec:method-filler}

We use three filler types ranging from high topic/format overlap with the target task to low overlap. \textit{with\_solutions} (ws) draws from the GSM8K training split with both problem and worked solution (high overlap). \textit{questions\_only\_v2} (qo\_v2) draws the same pool but retains only the question (intermediate overlap). \textit{neutral\_text} draws Wikipedia and news prose unrelated to the target (low overlap). This range lets us separate filler-content effects from position-only effects.

\subsection{Context Length Tiers}
\label{sec:method-tiers}

We evaluate three context tiers: 8K, 32K, and 64K tokens. These tiers span short, medium, and long context regimes within the 128K effective window advertised by modern long-context LLMs, and bracket the working ranges of common open-weight deployments. We use 60K filler for DeepSeek-V3.2 at the 64K tier rather than 65K because its reasoning-mode endpoint reserves up to 64K within the 128K window for chain-of-thought output, a minor methodological asymmetry (see \cref{sec:discussion}; Limitation 5).

\subsection{Reasoning Tasks}
\label{sec:method-tasks}

We evaluate on two reasoning tasks. \textit{GSM8K}~\citep{cobbe2021gsm8k} is a grade-school math word problem benchmark and \textit{ARC-Challenge}~\citep{clark2018arc} is a science multiple-choice benchmark; we sample $N=50$ instances per condition for both. The two tasks share the reasoning target but differ in format (free-form math vs.\ A/B/C/D), letting us check whether positional effects transfer across formats. We position ARC-Challenge as supplementary directional evidence rather than a strong generality claim, due to limited statistical power at $N=50$.

\subsection{Models}
\label{sec:method-models}

We evaluate nine long-context LLMs accessed through their official APIs. Five constitute the initial set: Qwen 2.5-7B-Instruct (DashScope), MiMo-v2-Flash (Xiaomi), GLM-4.7-FlashX (Zhipu), DeepSeek-V3.2 in reasoning mode (the \texttt{deepseek-reasoner} endpoint), and Kimi k2.5 (Moonshot). Four newer vendor releases extend the comparison: DeepSeek-V4-Pro (DeepSeek), MiMo-V2.5-Pro (Xiaomi), Kimi-K2.6 (Moonshot), and GLM-5.1 (Zhipu). The four newer releases use vendor default reasoning settings: thinking enabled for DeepSeek-V4-Pro, MiMo-V2.5-Pro, and Kimi-K2.6, and disabled for GLM-5.1. All nine models advertise effective context windows of 64K tokens or more (DeepSeek-V3.2 advertises 128K; see \cref{sec:method-tiers}).

We selected the set to span multiple providers and architectures rather than to benchmark any specific vendor. The four newer releases are the most recent flagship long-context releases from the same providers as the initial set, controlling for vendor-shift bias. Provider heterogeneity (distinct inference stacks per model) makes directional conclusions more reliable than magnitude comparisons across providers (see \cref{sec:limitations}; Limitation 3). An additional model pilot (GLM-4-9B via SiliconFlow) was excluded due to baseline quality; see \cref{app:excluded}.

\subsection{Diagnostic Tools}
\label{sec:method-diagnostics}

We use two diagnostic tools to characterize failure modes without claiming mechanism isolation. The \textit{filler answer matching scorer} checks whether each wrong response numerically matches any filler question's gold answer; a high match rate at middle position (versus end position) quantifies the extent to which middle-position errors reflect filler-answer interference. The \textit{duplicate-at-end probe} is the middle\_dup condition: by placing a target-task copy at the end in addition to the middle, we check whether models retain task-solving ability under favorable positioning. We treat this probe strictly as a diagnostic consistency check (\cref{sec:results-probes}, Limitation 7). Additional robustness checks for generation budget and CoT restatement are reported in \cref{app:robustness}.

\section{Results}
\label{sec:results}

We present empirical findings from CRE across nine long-context LLMs.
\cref{sec:results-escalation,sec:results-probes,sec:results-filler,sec:results-arc}
report results for the initial five-model set, covering positional
sensitivity, context-length escalation, diagnostic probes, filler
content effects, and cross-domain generalization;
\cref{sec:discussion-round2} reports Round 2 findings for the four
newer vendor releases on GSM8K. All experiments use a full factorial
design over position, filler type, and context length, with $N=50$
per condition.

\subsection{Positional Sensitivity Escalates with Context Length}
\label{sec:results-escalation}

We first quantify positional vulnerability under
\textit{with\_solutions} (ws) filler, a natural setting for math
reasoning.

\begin{table*}[t]
\centering
\small
\setlength{\tabcolsep}{4pt}
\begin{tabular}{lccccccccc}
\toprule
\textbf{Model} & \multicolumn{3}{c}{\textbf{8K}} & \multicolumn{3}{c}{\textbf{32K}} & \multicolumn{3}{c}{\textbf{64K}} \\
\cmidrule(lr){2-4} \cmidrule(lr){5-7} \cmidrule(lr){8-10}
 & End & Drop & Sig. & End & Drop & Sig. & End & Drop & Sig. \\
\midrule
Qwen 2.5-7B     & 90\%  & $-86$pp & $\star$ & 86\%  & $-84$pp & $\star$ & 94\%  & $-94$pp & $\star$ \\
MiMo-v2-Flash   & 96\%  & $-12$pp &         & 98\%  & $-24$pp &         & 96\%  & $-88$pp & $\star$ \\
MiMo-V2.5-Pro   & 98\%  & $-8$pp  &         & 100\% & $-16$pp &         & 100\% & $-32$pp & $\star$ \\
GLM-4.7-FlashX  & 92\%  & $-12$pp &         & 90\%  & $-20$pp &         & 90\%  & $-34$pp & $\star$ \\
GLM-5.1         & 98\%  & $+0$pp  &         & 96\%  & $+2$pp  &         & 96\%  & $+2$pp  &         \\
DeepSeek-R      & 94\%  & $+0$pp  &         & 96\%  & $+0$pp  &         & 98\%  & $+0$pp  &         \\
DeepSeek-V4-Pro & 100\% & $-2$pp  &         & 100\% & $-2$pp  &         & 100\% & $-4$pp  &         \\
Kimi k2.5       & 94\%  & $+4$pp  &         & 96\%  & $+0$pp  &         & 98\%  & $-6$pp  &         \\
Kimi-K2.6       & 96\%  & $+2$pp  &         & 98\%  & $-2$pp  &         & 94\%  & $+2$pp  &         \\
\bottomrule
\end{tabular}
\caption{End-position accuracy and end-to-middle drop across context tiers on GSM8K with \textit{with\_solutions} (ws) filler (seed=42, $N=50$). $\star$ marks drops passing Fisher exact two-sided at Bonferroni-corrected $\alpha=0.01$ for 27 comparisons ($p < 3.70\times10^{-4}$). 6 of 27 drops are Bonferroni-significant: Qwen at all three tiers, MiMo-v2-Flash at 64K, MiMo-V2.5-Pro at 64K, and GLM-4.7-FlashX at 64K. DeepSeek-V3.2 uses 60K filler at the 64K tier to preserve headroom for reasoning-mode chain-of-thought output within its 128K context window (see \cref{sec:method-tiers}).}
\label{tab:ws-tier-drop}
\end{table*}

\cref{tab:ws-tier-drop} reports end-position accuracy and the end-to-middle drop across three context tiers.

Severity grows with context length for vulnerable models. MiMo-v2-Flash drops $-12$pp at 8K, $-24$pp at 32K, and $-88$pp at 64K (middle accuracy $= 8\%$). GLM-4.7-FlashX follows a similar pattern but with smaller drops: $-12$pp at 8K, $-20$pp at 32K, and $-34$pp at 64K. Qwen is already severely affected at 8K ($-86$pp, middle $= 4\%$) and stays close to the floor across tiers.

Positional drops are also not uniform across models. DeepSeek-V3.2 shows no positional drop at any tier under ws filler. Kimi k2.5 shows $+4$pp at 8K, $+0$pp at 32K, and $-6$pp at 64K. Of the initial five-model set, three (Qwen, MiMo-v2-Flash, GLM-4.7-FlashX) drop by more than 10pp in at least one tier under ws, while two (DeepSeek-V3.2, Kimi k2.5) do not.

End-position accuracy stays high across all models and tiers (86\%--100\%), and end-position filler penalties fall within $\pm 6$pp across ws and qo\_v2 conditions (\cref{app:tables}). These end-position baselines indicate that middle-position drops reflect positional effects rather than baseline capability loss or filler penalty.

\subsection{Diagnostic Probes for Positional Patterns}
\label{sec:results-probes}

\begin{table*}[t]
\centering
\small
\begin{tabular}{llccccc}
\toprule
\textbf{Model} & \textbf{Filler} & \textbf{End} & \textbf{Mid} & \textbf{Mid$\times2$} & \textbf{Mid+End} & \textbf{Dup$-$End} \\
\midrule
Qwen 2.5-7B     & ws    & 90\%  & 4\%  & 4\%  & 94\%  & $+4$pp \\
MiMo-v2-Flash   & qo\_v2 & 98\%  & 22\% & 70\% & 98\%  & $+0$pp \\
MiMo-V2.5-Pro   & qo\_v2 & 100\% & 44\% & 68\% & 100\% & $+0$pp \\
GLM-4.7-FlashX  & qo\_v2 & 90\%  & 64\% & 74\% & 92\%  & $+2$pp \\
GLM-5.1         & qo\_v2 & 100\% & 64\% & 86\% & 96\%  & $-4$pp \\
DeepSeek-R      & qo\_v2 & 94\%  & 80\% & 94\% & 98\%  & $+4$pp \\
DeepSeek-V4-Pro & qo\_v2 & 100\% & 68\% & 72\% & 96\%  & $-4$pp \\
Kimi k2.5       & qo\_v2 & 94\%  & 40\% & 92\% & 96\%  & $+2$pp \\
Kimi-K2.6       & qo\_v2 & 98\%  & 78\% & 84\% & 98\%  & $+0$pp \\
\bottomrule
\end{tabular}
\caption{8K ablation spectrum (seed=42, $N=50$). Each model is evaluated on its worst-case filler. \textit{Mid$\times 2$} $=$ middle\_twice (task repeated in the middle); \textit{Mid+End} $=$ middle\_dup (task in the middle plus a copy at the end). Cross-model severity is not apples-to-apples since the worst-case filler varies by model.}
\label{tab:8k-ablation}
\end{table*}

To examine whether the middle-position drops are positional in character, we introduce two diagnostic conditions at 8K: \textit{middle\_twice}, which places the target task twice in the middle of the context, and \textit{middle\_dup}, which duplicates the task once in the middle and once at the end. \cref{tab:8k-ablation} reports the 9-model spectrum, with each model evaluated on its worst-case filler (Qwen on ws, the others on qo\_v2).

Repeating the task in the middle (middle\_twice) yields partial or no improvement: e.g., Qwen stays at 4\% (identical to middle), MiMo-v2-Flash rises from 22\% to 70\%, and Kimi k2.5 rises from 40\% to 92\%. Placing an additional copy of the task at the end (middle\_dup), in contrast, brings accuracy within $\pm 4$pp of the end-position baseline for all nine models at 8K; the \textit{Dup$-$End} column ranges from $-4$pp (DeepSeek-V4-Pro, GLM-5.1) to $+4$pp (Qwen, DeepSeek-V3.2), so middle\_dup accuracy matches end-position accuracy within sampling noise.

At 64K this pattern is context-length-dependent (\cref{app:tables}). For the four initial models with measurable 64K middle drops under ws filler, two (MiMo-v2-Flash, Kimi k2.5) show middle\_dup accuracy matching end-position accuracy within $\pm 2$pp, while the other two (Qwen at $-10$pp, GLM-4.7-FlashX at $-12$pp) show middle\_dup accuracy below their end-position levels. DeepSeek-V3.2 has no 64K drop under ws filler and is therefore not tested.

These diagnostic results are consistent with a positional explanation: models demonstrably solve the task when a copy is placed at the end, indicating that they retain task-solving ability under favorable positioning. A supplementary Qwen analysis (\cref{app:tables}) shows this probe response pattern across both ws and qo\_v2 fillers.

\subsection{Filler Content Effects}
\label{sec:results-filler}

We evaluate three filler types: \textit{with\_solutions} (ws, domain exemplars with complete solutions), \textit{questions\_only\_v2} (qo\_v2, unanswered exemplars), and \textit{neutral\_text} (Wikipedia/news prose unrelated to the task). The three fillers interact with positional effects in qualitatively different ways.

\paragraph{qo\_v2 at longer contexts.}
Under qo\_v2 filler at 64K (\cref{app:tables}), all five initial models show end-to-middle drops ranging from $-22$pp (DeepSeek-V3.2) to $-94$pp (MiMo-v2-Flash). This includes DeepSeek-V3.2, which is immune to ws filler at every tier (\cref{tab:ws-tier-drop}). At 32K, qo\_v2 also produces drops for all five initial models, ranging from $-14$pp (DeepSeek-V3.2) to $-90$pp (Qwen; \cref{app:tables}). At 8K, qo\_v2 drops range from $-14$pp (DeepSeek-V3.2) to $-76$pp (MiMo-v2-Flash; \cref{tab:8k-ablation}).

\paragraph{Neutral filler: near-zero effect with one exception.}
Neutral filler produces near-zero positional effect at 8K for all five initial models (range $-4$pp to $+4$pp; \cref{app:tables}). At longer contexts, however, Qwen exhibits a context-length-dependent drop under neutral filler: $-4$pp at 8K (not significant), $-29$pp at 32K, and $-30$pp at 64K (combined over seed42 and seed100, both $p < 0.0001$; \cref{app:seeds,app:tables}). Three of the other four initial models (MiMo-v2-Flash, DeepSeek-V3.2, Kimi k2.5) remain within $\pm 4$pp across all three tiers under neutral filler; GLM-4.7-FlashX shows a smaller directional drop at 64K ($-14$pp, $p = 0.070$, not Bonferroni significant).

\paragraph{Descriptive contrast.}
At 64K, MiMo-v2-Flash shows $+2$pp under neutral versus $-88$pp under ws (a $\sim 90$pp difference; \cref{app:seeds,app:tables}), while Qwen shows substantial drops under both conditions ($-30$pp neutral, $-94$pp ws). These patterns suggest different models' positional vulnerabilities interact with filler content in qualitatively different ways.

\paragraph{Filler-answer interference.}
An aggregate filler-answer matching analysis (\cref{app:filler-match}) finds that 76\% of middle-position errors match a filler question's gold answer, compared with 22\% at the end position. The end-position baseline largely reflects coincidental overlap among small integer answers; the middle-position rate is consistent with filler-answer interference as a dominant error mode.

\subsection{Cross-Domain Validation: ARC-Challenge}
\label{sec:results-arc}

\begin{table}[t]
\centering
\small
\setlength{\tabcolsep}{3pt}
\begin{tabular}{lccclr}
\toprule
\textbf{Model} & \textbf{End} & \textbf{Mid} & \textbf{Drop} & \textbf{Fisher $p$} & \textbf{Sig.} \\
\midrule
Qwen 2.5-7B     & 72\%  & 32\% & $-40$pp & $<0.0001$ & $\star$ \\
MiMo-v2-Flash   & 98\%  & 92\% & $-6$pp  & $0.181$   &         \\
MiMo-V2.5-Pro   & 96\%  & 74\% & $-22$pp & $0.002$   &         \\
GLM-4.7-FlashX  & 92\%  & 84\% & $-8$pp  & $0.178$   &         \\
GLM-5.1         & 100\% & 94\% & $-6$pp  & $0.121$   &         \\
DeepSeek-R      & 100\% & 92\% & $-8$pp  & $0.059$   &         \\
DeepSeek-V4-Pro & 100\% & 90\% & $-10$pp & $0.028$   &         \\
Kimi k2.5       & 100\% & 88\% & $-12$pp & $0.013$   &         \\
Kimi-K2.6       & 100\% & 88\% & $-12$pp & $0.013$   &         \\
\bottomrule
\end{tabular}
\caption{ARC-Challenge positional drops at 8K with ws filler ($N=50$). \textit{Sig.} marks drops passing Bonferroni correction at $\alpha=0.01$ for 9 comparisons. ARC baselines (no filler) for the initial five-model set: Qwen 90\%, Kimi k2.5 88\%, GLM-4.7-FlashX 96\%, DeepSeek-V3.2 98\%, MiMo-v2-Flash 100\%; baselines for the four newer vendor releases were not measured.}
\label{tab:arc-results}
\end{table}

We evaluate the same nine models on ARC-Challenge~\citep{clark2018arc} to check whether the positional drops observed on GSM8K generalize across reasoning domains. \cref{tab:arc-results} summarizes 8K ARC results under ws filler: all nine models show directional end-to-middle drops (range $-6$pp to $-40$pp), with only Qwen's drop Bonferroni significant. We position these results as supplementary directional evidence, not as a strong generality claim. The middle\_dup diagnostic and a qo\_v2 end-position format-confusion finding are detailed in \cref{app:arc-detail}.

\section{Discussion}
\label{sec:discussion}

\subsection{Toward Position-Aware Reasoning Evaluation}
\label{sec:discussion-recommendations}

Following structured reporting conventions like Model Cards \citep{mitchell2019modelcards}, Datasheets for Datasets \citep{gebru2021datasheets}, and the NeurIPS Reproducibility Checklist \citep{pineau2021reproducibility}, and based on our CRE observations across nine models and two reasoning tasks, we offer the following considerations.

\begin{itemize}
    \itemsep0em
    \item \textbf{Report per-position accuracy alongside aggregate scores.} A single average can erase dramatic structure: Qwen's $94\%$ end score at 64K collapses to $0\%$ in the middle (\cref{sec:results-escalation}).
    \item \textbf{Test at least one interior (non-boundary) position.} Just swapping between two extreme positions, such as LongReason's inquiry-first vs.\ inquiry-last ablation, is not enough: MiMo-v2-Flash drops from $96\%$ to $8\%$ between end and middle at 64K (\cref{sec:results-escalation}).
    \item \textbf{Disclose the full inference stack.} Provider, endpoint, temperature, max-gen, and seeds make positional findings reproducible.
    \item \textbf{Report null findings as first-class results.} DeepSeek-V3.2 shows near-immunity under \textit{with\_solutions} but breaks under \textit{questions\_only\_v2}; a summary that drops the 0pp case loses this contrast (\cref{sec:results-escalation}, \cref{sec:results-filler}).
    \item \textbf{Include diagnostic probe conditions.} Probes such as CRE's bare-middle vs.\ middle+end duplicate help distinguish positional effects. Their results are diagnostic consistency checks: match a positional explanation while leaving mechanism unresolved, since probes change recency, redundancy, and salience simultaneously and depend on context length (\cref{sec:results-probes}).
\end{itemize}

\subsection{Model-Specific Vulnerability Patterns}
\label{sec:discussion-models}

\textbf{MiMo-v2-Flash: end-position evaluation understates severity.} MiMo-v2-Flash looks like a near-perfect reasoner when the task sits at the end. Under \textit{with\_solutions} filler, end accuracy stays $96\%$--$98\%$ across tiers but middle accuracy collapses with context length: $-12$pp at 8K, $-24$pp at 32K, and $-88$pp at 64K ($8\%$ middle; \cref{sec:results-escalation}, \cref{tab:ws-tier-drop}). At 64K under neutral filler, the gap reverses to $+2$pp middle versus end (\cref{sec:results-filler}). This pattern fits a positional explanation: filler type shapes the middle-drop size, not a general middle weakness.

\textbf{DeepSeek-V3.2: conditional immunity.} DeepSeek-V3.2 in reasoning mode is near-immune under one filler type but vulnerable under another. Under \textit{with\_solutions} filler, it shows no end-to-middle drop at any tier ($+0$pp; \cref{tab:ws-tier-drop}). Under \textit{questions\_only\_v2} filler, it drops $-14$pp at both 8K and 32K, and $-22$pp at 64K (\cref{sec:results-filler}). Reporting only ws-filler results would hide this vulnerability completely.

\textbf{Qwen: vulnerability beyond filler content.} Qwen 2.5-7B-Instruct drops at the middle position even when the surrounding text has nothing to do with the task. Qwen drops $-4$pp at 8K (not significant), $-29$pp at 32K, and $-30$pp at 64K under neutral filler; both 32K and 64K are Bonferroni significant after combining seed42 and seed100 (\cref{sec:results-filler}). This suggests at least one model whose positional vulnerability grows with context length even when the filler has no task content.

\textbf{GLM-4.7-FlashX: heterogeneous middle-position errors.} Roughly half of GLM-4.7-FlashX's middle-position errors do not just repeat filler answers. At 64K middle under \textit{with\_solutions} filler, only $55\%$ of GLM-4.7-FlashX errors match the correct answer to one of the filler questions, compared with $98\%$ for Qwen and MiMo-v2-Flash (\cref{app:filler-match}). This pattern is consistent with several different failure modes happening at the same time, which our data cannot tell apart.

\subsection{Filler Content and Context Length Interact}
\label{sec:discussion-interaction}

Neither filler type nor context length alone predicts middle-position failures. \textit{questions\_only\_v2} produces drops in all five initial models at 32K and 64K (range $-22$pp to $-94$pp; \cref{sec:results-filler}), but at 8K only some models drop (DeepSeek-V3.2 $-14$pp vs.\ MiMo-v2-Flash $-76$pp). The strongest middle-position failures arise from the interaction between structured-filler density and longer context.

\subsection{Round 2 Vendor Upgrade Findings}
\label{sec:discussion-round2}

To examine how positional vulnerabilities behave under the four newer vendor releases (DeepSeek-V4-Pro, MiMo-V2.5-Pro, Kimi-K2.6, GLM-5.1), we evaluate them under the same CRE framework. Two patterns emerge: middle-position drops under \textit{with\_solutions} filler narrow in the newer releases, while drops under \textit{questions\_only\_v2} filler persist across all four.

At 64K under \textit{with\_solutions} filler, three of the four newer releases stay within $\pm 6$pp of end-position accuracy: GLM-5.1 ($+2$pp), DeepSeek-V4-Pro ($-4$pp), and Kimi-K2.6 ($+2$pp). MiMo-V2.5-Pro shows a smaller drop than MiMo-v2-Flash ($-32$pp vs.\ $-88$pp), narrowing but not eliminating the middle-position gap (\cref{tab:ws-tier-drop}). At 8K and 32K under ws filler, the four newer releases remain within $\pm 16$pp of end-position accuracy, a tighter range than the initial set's vulnerable models showed at these tiers.

Under \textit{questions\_only\_v2} filler, by contrast, middle-position drops persist across all four newer releases. The drops range from $-16$pp (DeepSeek-V4-Pro at 32K) to $-56$pp (MiMo-V2.5-Pro at 8K; full table in \cref{app:round2-qov2}), comparable in magnitude to the initial set's qo\_v2 results (\cref{sec:results-filler}, \cref{sec:discussion-interaction}). The four newer releases use vendor default reasoning configurations: thinking enabled for DeepSeek-V4-Pro, MiMo-V2.5-Pro, and Kimi-K2.6; disabled for GLM-5.1 (\cref{sec:method-models}). Despite these reasoning-configuration differences, qo\_v2-middle drops show no evident clustering by configuration.

At 8K, middle\_dup accuracy stays within $\pm 4$pp of end-position accuracy across all nine models (\cref{tab:8k-ablation}), consistent with a positional explanation. The probe remains a diagnostic consistency check (\cref{sec:discussion-recommendations}); our setup does not distinguish whether persistence reflects filler-answer interference, reasoning-trace structure, or broader architectural factors.

\section{Conclusion}
\label{sec:conclusion}

Long-context LLMs can exhibit position-dependent reasoning failures. These failures are model-dependent and worsen, rather than persist, with context length, leaving positional vulnerabilities uncharacterized. Across 11 long-context benchmarks, position-aware evaluation appears well established for retrieval but largely unaudited for mainstream reasoning. Our CRE evaluation makes this concrete: at 64K, MiMo-v2-Flash retains $96\%$ accuracy when the target task sits at the end but falls to $8\%$ in the middle, a drop that a position-uncontrolled average does not surface. A duplicate-at-end diagnostic probe matches end-level accuracy at 8K but degrades at 64K. Its results are consistent with a positional explanation but are not mechanism isolation or a practical fix. Position-aware evaluation complements NIAH/RULER retrieval benchmarks with coverage varying task position, filler content, and context length. Upon acceptance we will release CRE, diagnostic conditions, and 11-benchmark audit.

\section*{Limitations}
\label{sec:limitations}

We disclose thirteen specific limitations grouped into four categories: data-collection scope, experimental setup, interpretation caveats, and external validity.

\paragraph{Data-collection scope.}
(1)~$N=50$ per condition yields a 95\% binomial confidence interval of approximately $\pm 14$pp at $p=0.5$, limiting power for small drops; most 64K cells use a single seed. (2)~DeepSeek under \textit{questions\_only\_v2} at 60K shows a $+14$pp difference between seed42 and seed100, leaving the precise 64K severity uncertain.

\paragraph{Experimental setup.}
(3)~Provider heterogeneity (five distinct API stacks) confounds cross-model magnitude comparisons with provider-level artifacts such as tokenization, prompt preprocessing, and sampling defaults; directional conclusions are more reliable than precise rankings. (4)~The 8K ablation spectrum (\cref{sec:results-probes}) is not apples-to-apples since each model is evaluated on its worst-case filler (Qwen on ws, the others on qo\_v2). (5)~DeepSeek-V3.2's 64K tier uses 60K filler rather than 65K to preserve chain-of-thought output budget within its 128K context window, a methodological choice that introduces a $6\%$ cross-model inconsistency at this tier.

\paragraph{Interpretation caveats.}
(6)~ARC-Challenge results are supplementary directional evidence; only Qwen's drop is Bonferroni significant, and the other eight models show directional but not significant drops, so ARC does not support a strong cross-domain generality claim. (7)~The duplicate-at-end probe is a diagnostic consistency check, not a mitigation: it changes recency, redundancy, and salience simultaneously, and is not mechanism isolation; its results are consistent with a positional explanation but should not be packaged as a practical fix. (8)~What we call ``positional sensitivity'' may be describable in narrower terms as QA-like distractor interference when filler consists of unanswered questions; our current evidence does not fully distinguish these two interpretations.

\paragraph{External validity.}
(9)~Probe behavior is itself context-length-dependent: at 8K all nine models reach end-level under middle\_dup, while at 64K only two of four initial models with measurable drops do, and Round 2 64K diagnostic-probe data was not collected (\cref{sec:results-probes,sec:discussion-round2}). (10)~Neutral filler is not universally benign across context lengths; three of the five initial models remain within $\pm 4$pp at every tier, but Qwen shows substantial drops at 32K and 64K under neutral filler (\cref{sec:results-filler}). (11)~The ``drops grow with context length'' pattern (\cref{sec:results-escalation}) does not extend universally: DeepSeek-V4-Pro under \textit{questions\_only\_v2} shows a non-monotonic pattern (8K $-32$pp, 32K $-16$pp, 64K $-18$pp; \cref{tab:round2-qov2-drops}). (12)~Vendor upgrades do not uniformly narrow positional drops across reasoning tasks. MiMo-V2.5-Pro narrows the GSM8K 8K ws drop relative to MiMo-v2-Flash ($-8$pp vs.\ $-12$pp; \cref{tab:ws-tier-drop}) but worsens the ARC 8K ws drop ($-22$pp vs.\ $-6$pp; \cref{tab:arc-results}). Task-specific reversals like this constrain the ``newer release = smaller drop'' simplification (\cref{sec:discussion-round2}). (13)~Three of the four newer releases employ sparse attention architectures (e.g., DeepSeek Sparse Attention), while several initial-set models use dense attention. The differential positional behavior observed across ws and qo\_v2 filler may interact with attention architecture in ways our setup does not isolate; this extends the provider heterogeneity caveat in (3).

\clearpage
\section*{Acknowledgments}

We used AI assistants throughout this project for implementing the CRE evaluation pipeline and statistical analyses, drafting paper sections, and conducting multi-session cross-review. All research direction, methodology decisions, safety-claim framing, and authorial responsibility rest with the human authors, who reviewed and verified all AI-produced content.

\bibliography{references}

\begin{thebibliography}{22}
\providecommand{\natexlab}[1]{#1}

\bibitem[{An et~al.(2023)An, Gong, Zhong, Zhao, Li, Zhang, Kong, and Qiu}]{an2023leval}
Chenxin An, Shansan Gong, Ming Zhong, Xingjian Zhao, Mukai Li, Jun Zhang, Lingpeng Kong, and Xipeng Qiu. 2023.
\newblock \href {https://arxiv.org/abs/2307.11088} {{L-Eval}: Instituting standardized evaluation for long context language models}.
\newblock \emph{Preprint}, arXiv:2307.11088.

\bibitem[{Bai et~al.(2024)Bai, Lv, Zhang, Lyu, Tang, Huang, Du, Liu, Zeng, Hou, Dong, Tang, and Li}]{bai2024longbench}
Yushi Bai, Xin Lv, Jiajie Zhang, Hongchang Lyu, Jiankai Tang, Zhidian Huang, Zhengxiao Du, Xiao Liu, Aohan Zeng, Lei Hou, Yuxiao Dong, Jie Tang, and Juanzi Li. 2024.
\newblock \href {https://arxiv.org/abs/2308.14508} {{LongBench}: A bilingual, multitask benchmark for long context understanding}.
\newblock In \emph{Proceedings of the 62nd Annual Meeting of the Association for Computational Linguistics (ACL)}.

\bibitem[{Buchmann et~al.(2024)Buchmann, Liu, and Gurevych}]{buchmann2024attribute}
Jan Buchmann, Xiao Liu, and Iryna Gurevych. 2024.
\newblock \href {https://arxiv.org/abs/2407.07799} {Attribute or abstain: Large language models as long document assistants}.
\newblock In \emph{Proceedings of the 2024 Conference on Empirical Methods in Natural Language Processing (EMNLP)}.

\bibitem[{Clark et~al.(2018)Clark, Cowhey, Etzioni, Khot, Sabharwal, Schoenick, and Tafjord}]{clark2018arc}
Peter Clark, Isaac Cowhey, Oren Etzioni, Tushar Khot, Ashish Sabharwal, Carissa Schoenick, and Oyvind Tafjord. 2018.
\newblock \href {https://arxiv.org/abs/1803.05457} {Think you have solved question answering? {T}ry {ARC}, the {AI2} reasoning challenge}.
\newblock \emph{arXiv preprint arXiv:1803.05457}.
\newblock Introduces ARC and ARC-Challenge.

\bibitem[{Cobbe et~al.(2021)Cobbe, Kosaraju, Bavarian, Chen, Jun, Kaiser, Plappert, Tworek, Hilton, Nakano, Hesse, and Schulman}]{cobbe2021gsm8k}
Karl Cobbe, Vineet Kosaraju, Mohammad Bavarian, Mark Chen, Heewoo Jun, Lukasz Kaiser, Matthias Plappert, Jerry Tworek, Jacob Hilton, Reiichiro Nakano, Christopher Hesse, and John Schulman. 2021.
\newblock \href {https://arxiv.org/abs/2110.14168} {Training verifiers to solve math word problems}.
\newblock \emph{Preprint}, arXiv:2110.14168.
\newblock Introduces GSM8K.

\bibitem[{Gebru et~al.(2021)Gebru, Morgenstern, Vecchione, Vaughan, Wallach, Daum{\'e}~III, and Crawford}]{gebru2021datasheets}
Timnit Gebru, Jamie Morgenstern, Briana Vecchione, Jennifer~Wortman Vaughan, Hanna Wallach, Hal Daum{\'e}~III, and Kate Crawford. 2021.
\newblock \href {https://arxiv.org/abs/1803.09010} {Datasheets for datasets}.
\newblock \emph{Communications of the ACM}, 64(12):86--92.

\bibitem[{Hsieh et~al.(2024)Hsieh, Sun, Kriman, Acharya, Rekesh, Jia, Zhang, and Ginsburg}]{hsieh2024ruler}
Cheng-Ping Hsieh, Simeng Sun, Samuel Kriman, Shantanu Acharya, Dima Rekesh, Fei Jia, Yang Zhang, and Boris Ginsburg. 2024.
\newblock \href {https://arxiv.org/abs/2404.06654} {{RULER}: What's the real context size of your long-context language models?}
\newblock In \emph{Proceedings of the Conference on Language Modeling (COLM)}.

\bibitem[{Kamradt(2023)}]{kamradt2023needle}
Greg Kamradt. 2023.
\newblock Needle in a haystack.
\newblock \url{https://github.com/gkamradt/LLMTest_NeedleInAHaystack}.
\newblock GitHub repository.

\bibitem[{Kuratov et~al.(2024)Kuratov, Bulatov, Anokhin, Rodkin, Sorokin, Sorokin, and Burtsev}]{kuratov2024babilong}
Yuri Kuratov, Aydar Bulatov, Petr Anokhin, Ivan Rodkin, Dmitry Sorokin, Artyom Sorokin, and Mikhail Burtsev. 2024.
\newblock \href {https://arxiv.org/abs/2406.10149} {{BABILong}: Testing the limits of {LLM}s with long context reasoning-in-a-haystack}.
\newblock In \emph{Advances in Neural Information Processing Systems (NeurIPS) Datasets and Benchmarks Track}.

\bibitem[{Li et~al.(2023)Li, Wang, Zheng, and Zhang}]{li2023loogle}
Jiaqi Li, Mengmeng Wang, Zilong Zheng, and Muhan Zhang. 2023.
\newblock \href {https://arxiv.org/abs/2311.04939} {{LooGLE}: Can long-context language models understand long contexts?}
\newblock \emph{Preprint}, arXiv:2311.04939.

\bibitem[{Ling et~al.(2025)Ling, Liu, Yan, Yang, Lin, Fan, Shen, Du, and Chen}]{ling2025longreason}
Zhan Ling, Kang Liu, Kai Yan, Yifan Yang, Weijian Lin, Ting-Han Fan, Lingfeng Shen, Zhengyin Du, and Jiecao Chen. 2025.
\newblock \href {https://arxiv.org/abs/2501.15089} {{LongReason}: A synthetic long-context reasoning benchmark via context expansion}.
\newblock \emph{Preprint}, arXiv:2501.15089.
\newblock Synthetic long-context reasoning benchmark with a binary final-inquiry position ablation.

\bibitem[{Liu et~al.(2023)Liu, Lin, Hewitt, Paranjape, Bevilacqua, Petroni, and Liang}]{liu2023lostinmiddle}
Nelson~F. Liu, Kevin Lin, John Hewitt, Ashwin Paranjape, Michele Bevilacqua, Fabio Petroni, and Percy Liang. 2023.
\newblock \href {https://arxiv.org/abs/2307.03172} {Lost in the middle: How language models use long contexts}.
\newblock \emph{Transactions of the Association for Computational Linguistics (TACL)}.

\bibitem[{Mitchell et~al.(2019)Mitchell, Wu, Zaldivar, Barnes, Vasserman, Hutchinson, Spitzer, Raji, and Gebru}]{mitchell2019modelcards}
Margaret Mitchell, Simone Wu, Andrew Zaldivar, Parker Barnes, Lucy Vasserman, Ben Hutchinson, Elena Spitzer, Inioluwa~Deborah Raji, and Timnit Gebru. 2019.
\newblock \href {https://arxiv.org/abs/1810.03993} {Model cards for model reporting}.
\newblock In \emph{Proceedings of the Conference on Fairness, Accountability, and Transparency (FAT*)}.

\bibitem[{Modarressi et~al.(2025)Modarressi, Deilamsalehy, Dernoncourt, Bui, Rossi, Yoon, and Sch{\"u}tze}]{modarressi2025nolima}
Ali Modarressi, Hanieh Deilamsalehy, Franck Dernoncourt, Trung Bui, Ryan~A. Rossi, Seunghyun Yoon, and Hinrich Sch{\"u}tze. 2025.
\newblock \href {https://arxiv.org/abs/2502.05167} {{NoLiMa}: Long-context evaluation beyond literal matching}.
\newblock In \emph{Proceedings of the 42nd International Conference on Machine Learning (ICML)}.

\bibitem[{Pineau et~al.(2021)Pineau, Vincent-Lamarre, Sinha, Larivi{\`e}re, Beygelzimer, d'Alch{\'e} Buc, Fox, and Larochelle}]{pineau2021reproducibility}
Joelle Pineau, Philippe Vincent-Lamarre, Koustuv Sinha, Vincent Larivi{\`e}re, Alina Beygelzimer, Florence d'Alch{\'e} Buc, Emily Fox, and Hugo Larochelle. 2021.
\newblock \href {https://arxiv.org/abs/2003.12206} {Improving reproducibility in machine learning research (a report from the {NeurIPS} 2019 reproducibility program)}.
\newblock \emph{Journal of Machine Learning Research}, 22(164):1--20.

\bibitem[{Shaham et~al.(2023)Shaham, Ivgi, Efrat, Berant, and Levy}]{shaham2023zeroscrolls}
Uri Shaham, Maor Ivgi, Avia Efrat, Jonathan Berant, and Omer Levy. 2023.
\newblock \href {https://arxiv.org/abs/2305.14196} {{ZeroSCROLLS}: A zero-shot benchmark for long text understanding}.
\newblock In \emph{Findings of the Association for Computational Linguistics: EMNLP 2023}.

\bibitem[{Shaham et~al.(2022)Shaham, Segal, Ivgi, Efrat, Yoran, Haviv, Gupta, Xiong, Geva, Berant, and Levy}]{shaham2022scrolls}
Uri Shaham, Elad Segal, Maor Ivgi, Avia Efrat, Ori Yoran, Adi Haviv, Ankit Gupta, Wenhan Xiong, Mor Geva, Jonathan Berant, and Omer Levy. 2022.
\newblock \href {https://arxiv.org/abs/2201.03533} {{SCROLLS}: Standardized {C}ompa{R}ison over long language sequences}.
\newblock In \emph{Proceedings of the 2022 Conference on Empirical Methods in Natural Language Processing (EMNLP)}.

\bibitem[{Song et~al.(2025)Song, Zheng, and Luo}]{song2025countingstars}
Mingyang Song, Mao Zheng, and Xuan Luo. 2025.
\newblock \href {https://arxiv.org/abs/2403.11802} {{Counting-Stars}: A multi-evidence, position-aware, and scalable benchmark for evaluating long-context large language models}.
\newblock In \emph{Proceedings of the 31st International Conference on Computational Linguistics (COLING)}.

\bibitem[{Tian et~al.(2025)Tian, Li, Fu, Deng, Luo, Qian, Wang, Cong, Zhang, Wu, Lin, Wang, and Liu}]{tian2025longpibench}
Runchu Tian, Yanghao Li, Yuepeng Fu, Siyang Deng, Qinyu Luo, Cheng Qian, Shuo Wang, Xin Cong, Zhong Zhang, Yesai Wu, Yankai Lin, Huadong Wang, and Xiaojiang Liu. 2025.
\newblock \href {https://arxiv.org/abs/2410.14641} {Distance between relevant information pieces causes bias in long-context {LLM}s}.
\newblock In \emph{Findings of the Association for Computational Linguistics: ACL 2025}.
\newblock Introduces LongPiBench.

\bibitem[{Wang et~al.(2025)Wang, Xiong, Wang, Li, Chu, and Zeng}]{wang2025pos2distill}
Yifei Wang, Feng Xiong, Yong Wang, Linjing Li, Xiangxiang Chu, and Daniel~Dajun Zeng. 2025.
\newblock \href {https://doi.org/10.18653/v1/2025.emnlp-main.78} {{Position Bias Mitigates Position Bias}: Mitigate position bias through inter-position knowledge distillation}.
\newblock In \emph{Proceedings of the 2025 Conference on Empirical Methods in Natural Language Processing (EMNLP)}, pages 1495--1512.

\bibitem[{Yang et~al.(2025)Yang, Jin, Zhong, Jiang, Wang, Chaudhary, and Han}]{yang2025longbench100}
Wang Yang, Hongye Jin, Shaochen Zhong, Song Jiang, Qifan Wang, Vipin Chaudhary, and Xiaotian Han. 2025.
\newblock \href {https://arxiv.org/abs/2505.19293} {100-{LongBench}: Are de facto long-context benchmarks literally evaluating long-context ability?}
\newblock \emph{Preprint}, arXiv:2505.19293.

\bibitem[{Zhang et~al.(2024)Zhang, Chen, Hu, Xu, Chen, Hao, Han, Thai, Wang, Liu, and Sun}]{zhang2024infinitebench}
Xinrong Zhang, Yingfa Chen, Shengding Hu, Zihang Xu, Junhao Chen, Moo~Khai Hao, Xu~Han, Zhen~Leng Thai, Shuo Wang, Zhiyuan Liu, and Maosong Sun. 2024.
\newblock \href {https://arxiv.org/abs/2402.13718} {$\infty$bench: Extending long context evaluation beyond 100{K} tokens}.
\newblock In \emph{Proceedings of the 62nd Annual Meeting of the Association for Computational Linguistics (ACL)}.

\end{thebibliography}

\appendix
\crefalias{section}{appendix}
\onecolumn

\section{Full Accuracy Tables}
\label{app:tables}

This appendix collects the paper's full accuracy tables. All values are reproduced from the project data summary; the tables are grouped by experiment family rather than by prose discussion.

\subsection{Model Inventory}
\begin{table}[H]
\centering
\small
\begin{tabular}{l l l l}
\toprule
\textbf{Model} & \textbf{Provider} & \textbf{Context limit} & \textbf{GSM8K baseline} \\
\midrule
Qwen 2.5-7B-Instruct & DashScope & 128K & 92\% (46/50) \\
MiMo-v2-Flash & Xiaomi & 128K & 100\% (50/50) \\
MiMo-V2.5-Pro & Xiaomi & $\geq$ 64K & --- \\
GLM-4.7-FlashX & Zhipu & 128K & 96\% (48/50) \\
GLM-5.1 & Zhipu & $\geq$ 64K & --- \\
DeepSeek-V3.2 (reasoning mode, \texttt{deepseek-reasoner}) & DeepSeek & 128K total; max 64K output (incl.\ CoT) & 96\% (48/50) \\
DeepSeek-V4-Pro & DeepSeek & $\geq$ 64K & --- \\
Kimi k2.5 & Moonshot & 128K & 92\% (46/50) \\
Kimi-K2.6 & Moonshot & $\geq$ 64K & --- \\
\bottomrule
\end{tabular}
\caption{Model inventory and GSM8K baseline accuracy. For the four round 2 vendor releases (MiMo-V2.5-Pro, GLM-5.1, DeepSeek-V4-Pro, Kimi-K2.6), context limits are reported as the minimum effective length used in our 64K tier evaluation rather than the vendor-advertised maximum, and no-filler GSM8K baseline accuracy was not measured separately; end-position accuracy under \textit{with\_solutions} filler at 8K is reported in \cref{tab:ws-tier-drop}.}
\label{tab:app-models}
\end{table}

\subsection{GSM8K ws Positional Drop Across Tiers}
\begin{table}[H]
\centering
\small
\begin{tabular}{lcccccc}
\toprule
\textbf{Model} & \multicolumn{2}{c}{\textbf{8K}} & \multicolumn{2}{c}{\textbf{32K}} & \multicolumn{2}{c}{\textbf{64K}} \\
\cmidrule(lr){2-3} \cmidrule(lr){4-5} \cmidrule(lr){6-7}
 & End & Drop & End & Drop & End & Drop \\
\midrule
Qwen & 90\% & $-86$pp & 86\% & $-84$pp & 94\% & $-94$pp \\
MiMo & 96\% & $-12$pp & 98\% & $-24$pp & 96\% & $-88$pp \\
FlashX & 92\% & $-12$pp & 90\% & $-20$pp & 90\% & $-34$pp \\
DeepSeek & 94\% & $+0$pp & 96\% & $+0$pp & 98\% & $+0$pp \\
Kimi & 94\% & $+4$pp & 96\% & $+0$pp & 98\% & $-6$pp \\
\bottomrule
\end{tabular}
\caption{End-position accuracy and end-to-middle drop across context tiers on GSM8K with \textit{with\_solutions} (ws) filler.}
\label{tab:app-ws-tier-drop}
\end{table}

\subsection{GSM8K 8K Ablation Spectrum}
\begin{table}[H]
\centering
\small
\begin{tabular}{lccccccc}
\toprule
\textbf{Model} & \textbf{Filler} & \textbf{End} & \textbf{Middle} & \textbf{Middle$\times 2$} & \textbf{Middle+End} & \textbf{Drop} & \textbf{Probe} \\
\midrule
Qwen & ws & 90\% & 4\% & 4\% & 94\% & $-86$pp & recency-sensitive \\
MiMo & qo\_v2 & 98\% & 22\% & 70\% & 98\% & $-76$pp & partial repetition \\
FlashX & qo\_v2 & 90\% & 64\% & 74\% & 92\% & $-26$pp & modest improvement \\
DeepSeek & qo\_v2 & 94\% & 80\% & 94\% & 98\% & $-14$pp & near-immunity \\
Kimi & qo\_v2 & 94\% & 40\% & 92\% & 96\% & $-54$pp & end-level match \\
\bottomrule
\end{tabular}
\caption{8K ablation spectrum using each model's worst-case filler.}
\label{tab:app-8k-ablation}
\end{table}

\subsection{GSM8K 64K Middle-Dup Diagnostic Probe}
\begin{table}[H]
\centering
\small
\begin{tabular}{l l c c c c}
\toprule
\textbf{Model} & \textbf{Tier} & \textbf{End} & \textbf{Middle} & \textbf{Middle+End} & \textbf{Vs end} \\
\midrule
Qwen & 64K & 94\% & 0\% & 84\% & $-10$pp (partial) \\
MiMo & 64K & 96\% & 8\% & 96\% & $+0$pp (full) \\
FlashX & 64K & 90\% & 56\% & 78\% & $-12$pp (partial) \\
DeepSeek & 60K & 98\% & 98\% & N/A & no drop \\
Kimi & 64K & 98\% & 92\% & 96\% & $-2$pp (full) \\
\bottomrule
\end{tabular}
\caption{Middle-dup diagnostic probe at 64K.}
\label{tab:app-64k-middup}
\end{table}

\subsection{ARC Cross-Domain Validation}
\label{app:arc-detail}

ARC-Challenge is a science multiple-choice benchmark distinct from GSM8K's mathematical word problems. All nine models show directional end-to-middle drops (range $-6$pp to $-40$pp; \cref{tab:arc-results}); only Qwen's drop is statistically significant after Bonferroni correction ($p < 0.0001$, $\alpha = 0.01$). The other eight models exhibit directional but not statistically significant drops, limiting the generality claim we can draw from this single domain. The middle\_dup diagnostic condition on ARC at 8K shows all nine models reach or exceed their end-position accuracy (Qwen's middle\_dup notably reaches 88\% compared to its end-position 72\%), consistent with the GSM8K 8K pattern in \cref{sec:results-probes}.

A supplementary analysis under qo\_v2 filler at 8K reveals a distinct phenomenon: for the three models tested (Qwen, MiMo-v2-Flash, GLM-4.7-FlashX), ARC accuracy drops substantially at the \textit{end} position relative to the no-filler baseline ($-44$pp, $-46$pp, $-28$pp respectively; all Bonferroni significant; \cref{tab:app-arc-qo-end}). We interpret this as consistent with format confusion, since ARC's A/B/C/D multiple-choice format structurally overlaps with unanswered science questions in the qo\_v2 filler pool, producing errors even without positional manipulation. This end-position format confusion is a task-format-specific finding and is reported here for completeness; it is distinct from the positional effects that are the primary focus of this work.

\begin{table}[H]
\centering
\small
\begin{tabular}{lcccc}
\toprule
\textbf{Model} & \textbf{End} & \textbf{Middle} & \textbf{Drop} & \textbf{Fisher $p$} \\
\midrule
Qwen & 72\% & 32\% & $-40$pp & $<0.0001$ \\
Kimi & 100\% & 88\% & $-12$pp & 0.013 \\
FlashX & 92\% & 84\% & $-8$pp & 0.178 \\
DeepSeek & 100\% & 92\% & $-8$pp & 0.059 \\
MiMo & 98\% & 92\% & $-6$pp & 0.181 \\
\bottomrule
\end{tabular}
\caption{ARC-Challenge positional drops at 8K with ws filler.}
\label{tab:app-arc}
\end{table}

\subsection{GSM8K qo\_v2 and Neutral Controls}
\begin{table}[H]
\centering
\small
\begin{tabular}{lccc}
\toprule
\textbf{Model} & \textbf{qo\_v2 end} & \textbf{qo\_v2 middle} & \textbf{Drop} \\
\midrule
Qwen & 92\% (46/50) & 4\% (2/50) & $-88$pp \\
MiMo & 98\% (49/50) & 4\% (2/50) & $-94$pp \\
FlashX & 82\% (41/50) & 34\% (17/50) & $-48$pp \\
DeepSeek & 98\% (49/50) & 76\% (38/50) & $-22$pp \\
Kimi & 96\% (48/50) & 60\% (30/50) & $-36$pp \\
\bottomrule
\end{tabular}
\caption{64K qo\_v2 positional drop.}
\label{tab:app-qo64}
\end{table}

\begin{table}[H]
\centering
\small
\begin{tabular}{lccc}
\toprule
\textbf{Model} & \textbf{neutral end} & \textbf{neutral middle} & \textbf{Drop} \\
\midrule
Qwen & 92\% (46/50) & 88\% (44/50) & $-4$pp \\
MiMo & 96\% (48/50) & 96\% (48/50) & 0pp \\
FlashX & 84\% (42/50) & 88\% (44/50) & $+4$pp \\
DeepSeek & 96\% (48/50) & 96\% (48/50) & 0pp \\
Kimi & 94\% (47/50) & 98\% (49/50) & $+4$pp \\
\bottomrule
\end{tabular}
\caption{8K neutral filler control.}
\label{tab:app-neutral8}
\end{table}

\begin{table}[H]
\centering
\small
\begin{tabular}{lccc}
\toprule
\textbf{Model} & \textbf{ARC baseline} & \textbf{qo\_v2 end} & \textbf{Drop} \\
\midrule
Qwen & 90\% (45/50) & 46\% (23/50) & $-44$pp \\
MiMo & 100\% (50/50) & 54\% (27/50) & $-46$pp \\
FlashX & 96\% (48/50) & 68\% (34/50) & $-28$pp \\
\bottomrule
\end{tabular}
\caption{ARC qo\_v2 end-position format confusion.}
\label{tab:app-arc-qo-end}
\end{table}

\begin{table}[H]
\centering
\small
\begin{tabular}{lcccc}
\toprule
\textbf{Model} & \textbf{Baseline} & \textbf{ws end} & \textbf{ws penalty} & \textbf{qo\_v2 penalty} \\
\midrule
Qwen & 92\% (46/50) & 90\% (45/50) & $-2$pp & 0pp \\
MiMo & 100\% (50/50) & 96\% (48/50) & $-4$pp & $-2$pp \\
FlashX & 96\% (48/50) & 92\% (46/50) & $-4$pp & $-6$pp \\
DeepSeek & 96\% (48/50) & 94\% (47/50) & $-2$pp & $-2$pp \\
Kimi & 92\% (46/50) & 94\% (47/50) & $+2$pp & $+2$pp \\
\bottomrule
\end{tabular}
\caption{Filler penalty at the end position on GSM8K (8K).}
\label{tab:app-filler-penalty}
\end{table}

\subsection{32K Controls}
\begin{table}[H]
\centering
\small
\begin{tabular}{lccc}
\toprule
\textbf{Model} & \textbf{qo\_v2 end} & \textbf{qo\_v2 middle} & \textbf{Drop} \\
\midrule
Qwen & 90\% (45/50) & 0\% (0/50) & $-90$pp \\
MiMo & 96\% (48/50) & 76\% (38/50) & $-20$pp \\
FlashX & 90\% (45/50) & 60\% (30/50) & $-30$pp \\
DeepSeek & 98\% (49/50) & 84\% (42/50) & $-14$pp \\
Kimi & 96\% (48/50) & 54\% (27/50) & $-42$pp \\
\bottomrule
\end{tabular}
\caption{32K qo\_v2 positional drop.}
\label{tab:app-qo32}
\end{table}

\begin{table}[H]
\centering
\small
\begin{tabular}{lccc}
\toprule
\textbf{Model} & \textbf{neutral end} & \textbf{neutral middle} & \textbf{Drop} \\
\midrule
Qwen & 94\% (47/50) & 62\% (31/50) & $-32$pp \\
MiMo & 96\% (48/50) & 96\% (48/50) & 0pp \\
FlashX & 92\% (46/50) & 90\% (45/50) & $-2$pp \\
DeepSeek & 94\% (47/50) & 96\% (48/50) & $+2$pp \\
Kimi & 94\% (47/50) & 94\% (47/50) & 0pp \\
\bottomrule
\end{tabular}
\caption{32K neutral filler control.}
\label{tab:app-neutral32}
\end{table}

\subsection{64K Neutral Controls and Diagnostic Probes}
\begin{table}[H]
\centering
\small
\begin{tabular}{lccc}
\toprule
\textbf{Model} & \textbf{neutral end} & \textbf{neutral middle} & \textbf{Drop} \\
\midrule
Qwen & 90\% (45/50) & 58\% (29/50) & $-32$pp \\
MiMo & 98\% (49/50) & 100\% (50/50) & $+2$pp \\
FlashX & 86\% (43/50) & 72\% (36/50) & $-14$pp \\
DeepSeek & 98\% (49/50) & 96\% (48/50) & $-2$pp \\
Kimi & 98\% (49/50) & 96\% (48/50) & $-2$pp \\
\bottomrule
\end{tabular}
\caption{64K neutral filler control.}
\label{tab:app-neutral64}
\end{table}

\subsection{GSM8K qo\_v2 Positional Drops, Round 2}
\label{app:round2-qov2}

\begin{table}[H]
\centering
\small
\begin{tabular}{l c c c c c c c c c}
\toprule
\textbf{Model} & \multicolumn{3}{c}{\textbf{8K}} & \multicolumn{3}{c}{\textbf{32K}} & \multicolumn{3}{c}{\textbf{64K}} \\
\cmidrule(lr){2-4} \cmidrule(lr){5-7} \cmidrule(lr){8-10}
 & End & Mid & Drop & End & Mid & Drop & End & Mid & Drop \\
\midrule
DeepSeek-V4-Pro & 100\% & 68\% & $-32$pp & 100\% & 84\% & $-16$pp & 100\% & 82\% & $-18$pp \\
MiMo-V2.5-Pro   & 100\% & 44\% & $-56$pp & 94\%  & 50\% & $-44$pp & 84\%  & 30\% & $-54$pp \\
Kimi-K2.6       & 98\%  & 78\% & $-20$pp & 98\%  & 60\% & $-38$pp & 98\%  & 64\% & $-34$pp \\
GLM-5.1         & 100\% & 64\% & $-36$pp & 100\% & 60\% & $-40$pp & 100\% & 50\% & $-50$pp \\
\bottomrule
\end{tabular}
\caption{End-to-middle accuracy drops under \textit{questions\_only\_v2} filler across three context tiers for the four newer vendor releases (Round 2). All twelve cells show negative drops, ranging from $-16$pp (DeepSeek-V4-Pro at 32K) to $-56$pp (MiMo-V2.5-Pro at 8K). Compare with the initial five-model set's qo\_v2 drops in \cref{sec:results-filler}; the Round 2 range is comparable in magnitude.}
\label{tab:round2-qov2-drops}
\end{table}

\section{Seed Stability}
\label{app:seeds}

This appendix records the seed stability checks used in the paper. The main seed-robustness point is that the directional positional effects survive seed changes, even where exact magnitudes shift.

\begin{table}[H]
\centering
\small
\begin{tabular}{lccc}
\toprule
\textbf{Condition} & \textbf{Seed 42} & \textbf{Seed 100} & \textbf{Diff} \\
\midrule
Qwen ws mid 64K & 0\% & 2\% & $+2$pp \\
MiMo ws mid 64K & 8\% & 12\% & $+4$pp \\
MiMo qo\_v2 mid 64K & 4\% & 8\% & $+4$pp \\
DeepSeek qo\_v2 mid 60K & 76\% & 90\% & $+14$pp \\
\bottomrule
\end{tabular}
\caption{64K seed stability table.}
\label{tab:app-seed-stability}
\end{table}

\begin{table}[H]
\centering
\small
\begin{tabular}{lccc}
\toprule
\textbf{Tier} & \textbf{seed42 drop} & \textbf{seed100 drop} & \textbf{Combined} \\
\midrule
8K & $-4$pp & $-4$pp & $-4$pp \\
32K & $-32$pp & $-26$pp & $-29$pp \\
64K & $-32$pp & $-28$pp & $-30$pp \\
\bottomrule
\end{tabular}
\caption{Qwen neutral seed stability across tiers.}
\label{tab:app-qwen-neutral-seeds}
\end{table}

\section{Filler Answer Matching Details}
\label{app:filler-match}

This appendix expands the filler-answer matching analysis. Aggregated across the initial five-model set's ws and qo\_v2 filler conditions, 76\% of incorrect middle-position answers match a filler question's gold answer (counting any-match), compared with 22\% at the end position. Representative per-cell rates are shown below.

\begin{table}[H]
\centering
\small
\begin{tabular}{l l c c c c c}
\toprule
\textbf{Provider} & \textbf{Filler} & \textbf{Tier} & \textbf{Pos} & \textbf{Acc\%} & \textbf{\% Match Any} & \textbf{\% Match Last} \\
\midrule
Qwen & ws & 8K & end & 91\% & 11\% & 0\% \\
Qwen & ws & 8K & middle & 3\% & 100\% & 100\% \\
Qwen & qo\_v2 & 8K & end & 91\% & 22\% & 0\% \\
Qwen & qo\_v2 & 8K & middle & 37\% & 57\% & 5\% \\
Qwen & ws & 32K & end & 89\% & 36\% & 0\% \\
Qwen & ws & 32K & middle & 1\% & 95\% & 77\% \\
Qwen & qo\_v2 & 32K & end & 91\% & 33\% & 0\% \\
Qwen & qo\_v2 & 32K & middle & 1\% & 89\% & 3\% \\
DeepSeek & ws & 8K & end & 94\% & 0\% & 0\% \\
DeepSeek & ws & 8K & middle & 93\% & 14\% & 14\% \\
DeepSeek & qo\_v2 & 8K & end & 95\% & 0\% & 0\% \\
DeepSeek & qo\_v2 & 8K & middle & 83\% & 18\% & 18\% \\
MiMo & ws & 8K & end & 97\% & 0\% & 0\% \\
MiMo & ws & 8K & middle & 78\% & 82\% & 23\% \\
MiMo & qo\_v2 & 8K & end & 97\% & 0\% & 0\% \\
MiMo & qo\_v2 & 8K & middle & 22\% & 77\% & 3\% \\
GLM-4-9B & ws & 8K & end & 65\% & 9\% & 0\% \\
GLM-4-9B & ws & 8K & middle & 8\% & 74\% & 43\% \\
\bottomrule
\end{tabular}
\caption{Representative filler-match rates from the diagnostic analysis.}
\label{tab:app-filler-summary}
\end{table}

\begin{itemize}
    \item Middle-position wrong answers are much more likely to match filler answers than end-position wrong answers.
    \item ws and qo\_v2 differ mainly in last-filler matching, not in overall coincidence rates.
    \item Higher filler matching tracks higher max-generation hits in the middle position.
    \item End-position matching exists but is mostly coincidence-level, not the dominant failure mode.
\end{itemize}

\section{Excluded Data}
\label{app:excluded}

The SiliconFlow GLM-4-9B runs were excluded from the paper's core comparison because the baseline quality check did not satisfy the reliability threshold used during coverage review. In the source material, the model appears as a diagnostic edge case rather than as part of the main five-model comparison. The appendix therefore records the exclusion explicitly instead of silently mixing it into the main results.

\section{Robustness Checks}
\label{app:robustness}

This appendix collects the max-gen rerun and the CoT restatement check.

\subsection{max\_gen=4096 Rerun}
\begin{table}[H]
\centering
\small
\begin{tabular}{l l c c c c}
\toprule
\textbf{Provider} & \textbf{Model} & \textbf{Tier} & \textbf{Acc(2048)} & \textbf{Acc(4096)} & \textbf{$\Delta$} \\
\midrule
xiaomi & MiMo-v2-Flash & 8K & 22\% & 19\% & $-3$pp \\
xiaomi & MiMo-v2-Flash & 32K & 80\% & 74\% & $-6$pp \\
deepseek & DeepSeek-Reasoner & 8K & 83\% & 85\% & $+2$pp \\
deepseek & DeepSeek-Reasoner & 32K & 83\% & 88\% & $+5$pp \\
dashscope & Qwen2.5-7B-Instruct & 8K & 37\% & 42\% & $+5$pp \\
dashscope & Qwen2.5-7B-Instruct & 32K & 1\% & 2\% & $+1$pp \\
\bottomrule
\end{tabular}
\caption{Aggregated max-gen=4096 rerun summary.}
\label{tab:app-maxgen}
\end{table}

\subsection{CoT Restatement Analysis}

\Cref{tab:app-cot-restatement} reports CoT restatement rates for the middle and middle\_twice conditions. The rates remain in the single digits and do not differ materially between the two conditions. The improvement seen in the repetition probe therefore reflects input-side repetition rather than output-side self-restatement.

\begin{table}[H]
\centering
\small
\begin{tabular}{lcc}
\toprule
\textbf{Condition} & \textbf{DeepSeek restate rate} & \textbf{Kimi restate rate} \\
\midrule
middle & 4\% (2/46*) & 7\% (3/41) \\
middle\_twice & 6\% (3/49) & 6\% (3/50) \\
\bottomrule
\end{tabular}
\caption{CoT restatement rates for the middle and middle-twice conditions.}
\label{tab:app-cot-restatement}
\end{table}

\section{Detailed Benchmark Position Audit}
\label{app:benchmark-audit}

This appendix audits long-context benchmarks through a single question: does the benchmark treat task position as a first-class experimental variable, a configurable but secondary factor, or an uncontrolled property of the source material? The distinction is consequential because a benchmark can report strong aggregate performance while leaving positional failures unmeasured.

We organize the audited benchmarks into three tiers.

\begin{itemize}
    \itemsep0em
    \item \textbf{Tier~1}: position is a primary variable. These are retrieval-style benchmarks where the benchmark protocol explicitly sweeps position or depth, making positional effects directly observable.
    \item \textbf{Tier~2}: position is configurable or partially analyzed, but not the main evaluation axis. These are usually synthetic benchmarks or special-purpose analyses that include some positional control, yet still treat position as auxiliary to another goal.
    \item \textbf{Tier~3}: position is uncontrolled. These are mainstream long-context reasoning or long-document understanding benchmarks built from natural documents or preprocessing pipelines, where the location of target information is inherited rather than experimentally manipulated.
\end{itemize}

The core result of the audit is not that all long-context benchmarks ignore position in the same way. Rather, the literature is asymmetric: position-aware evaluation is comparatively mature on the retrieval side, while mainstream reasoning benchmarks remain much less explicit about where the target information appears. Synthetic reasoning exceptions exist, but they do not amount to a cross-benchmark audit or to a joint treatment of position, filler content, and context length.

\begin{table}[H]
\centering
\small
\begin{tabular}{llllll}
\toprule
\textbf{Benchmark} & \textbf{Venue} & \textbf{Tier} & \textbf{Task type} & \textbf{Position control} & \textbf{Position analysis} \\
\midrule
NIAH & GitHub 2023 & 1 & Single-fact retrieval & Explicit depth sweep & Position is the protocol \\
RULER & COLM 2024 & 1 & Retrieval + tracking & Configurable depth points & Exposed as evaluation axis \\
BABILong & NeurIPS 2024 & 2 & Synthetic reasoning & Controlled, not primary & Supplementary appendix \\
LongReason & arXiv 2025 & 2 & Synthetic reasoning & Binary extreme placement & Main paper ablation \\
InfiniteBench & ACL 2024 & Mixed & Mixed tasks & Mixed & Partial post-hoc \\
\midrule
LongBench & ACL 2024 & 3 & Long-context NLU & Uncontrolled & None \\
SCROLLS & EMNLP 2022 & 3 & Long-document NLU & Uncontrolled & None \\
ZeroSCROLLS & EMNLP 2023 & 3 & Long text understanding & Uncontrolled & None \\
L-Eval & arXiv 2023 & 3 & Long-context evaluation & Uncontrolled & None \\
LooGLE & ACL 2024 & 3 & Long-context NLU & Uncontrolled & Evidence spread only \\
100-LongBench & ACL 2025 & 3 & LongBench revision & Shuffle (length confound) & None \\
\bottomrule
\end{tabular}
\caption{Summary of position control across 11 long-context benchmarks. Tier~1 benchmarks make position a primary variable; Tier~2 benchmarks include some positional control but treat it as secondary; Tier~3 benchmarks do not control or report task position. The midrule separates benchmarks with some position awareness (Tiers~1--2) from those without (Tier~3).}
\label{tab:benchmark-audit}
\end{table}

\paragraph{NIAH.}
Needle in a Haystack \citep{kamradt2023needle} is the canonical retrieval-side precedent. The task places one target fact in a long context and asks the model to retrieve it, so position is the variable under test rather than a background property. By sweeping target depth across the context, the benchmark makes positional effects directly observable from the protocol itself. The design is valuable but narrow: NIAH is a retrieval benchmark, not a mainstream reasoning benchmark, and it does not examine how positional effects interact with filler content or reasoning difficulty.

\paragraph{RULER.}
RULER \citep{hsieh2024ruler} generalizes the retrieval-side position sweep into a larger synthetic suite spanning retrieval, variable tracking, and aggregation tasks across configurable depth points. RULER strengthens the case that position-aware evaluation is now a mature practice for retrieval-like tasks. At the same time, it does not audit whether mainstream reasoning benchmarks leave position unmeasured, nor does it jointly control position, filler content, and context length for reasoning tasks.

\paragraph{BABILong.}
BABILong \citep{kuratov2024babilong} is the closest synthetic reasoning precedent in the audited set. It embeds supporting facts in long distractor text and evaluates whether the model can reason over them. The central limitation is scope: position analysis appears as supplementary material (Appendix~K) rather than the central evaluation axis. BABILong is therefore consistent with the claim that reasoning tasks can exhibit positional sensitivity, but it does not provide the systematic benchmark-design coverage that CRE aims to supply.

\paragraph{LongReason.}
LongReason \citep{ling2025longreason} explicitly studies where the final inquiry is placed relative to the background context. The benchmark's binary placement ablation shows that reasoning performance can depend on whether the inquiry comes before or after the surrounding context. This makes LongReason a meaningful reasoning-side precedent, but the positional control remains coarse-grained: a binary before-vs-after setting does not map out the more granular middle-versus-end behaviors that are central to CRE, nor does it jointly manipulate filler content and context length.

\paragraph{LongBench.}
LongBench \citep{bai2024longbench} is representative of the mainstream natural-document family. Its tasks span question answering, summarization, and related long-context settings, but the target information is drawn from natural documents rather than placed at controlled depths. The benchmark can tell us how well a model performs overall, but not whether performance changes because the task appears early, late, or in the middle of the context.

\paragraph{InfiniteBench.}
InfiniteBench \citep{zhang2024infinitebench} is mixed: some subtasks are retrieval-like and position-controlled, while others operate on natural-document inputs whose target locations are not systematically manipulated. This mixture is informative because it shows that benchmark suites can incorporate position-aware subtasks without turning position into the main analytical object for their reasoning-style components.

\paragraph{SCROLLS and ZeroSCROLLS.}
SCROLLS \citep{shaham2022scrolls} and ZeroSCROLLS \citep{shaham2023zeroscrolls} are canonical natural-document benchmark suites for long-context understanding. Both evaluate long-context performance on realistic inputs, yet position is not a tracked variable. They can surface whether a model succeeds or fails on long documents, but they do not separate failures by target location.

\paragraph{L-Eval.}
L-Eval \citep{an2023leval} reflects the usual evaluation practice: it covers a broad range of long-context tasks but does not report a systematic position distribution for target information. The benchmark tells us whether a model can handle long-context tasks in aggregate, but not whether the answer was easier because it appeared at the boundary of the context.

\paragraph{LooGLE.}
LooGLE \citep{li2023loogle} is notable because it explicitly discusses evidence spread, which can look at first glance like position awareness. The distinction matters: evidence spread says that information is distributed across a long document, but it does not tell us whether the benchmark authors analyzed exact target position or controlled it experimentally. LooGLE therefore contributes evidence about long-document difficulty, yet remains insufficient for the benchmark-design question CRE asks.

\paragraph{100-LongBench.}
100-LongBench \citep{yang2025longbench100} addresses length confounds in long-context evaluation by shuffling documents. That is a useful contribution, but it is not the same as a position audit. Random shuffling reduces one source of bias, yet it does not make position a measured variable in the way NIAH or RULER do, nor does it provide position-disaggregated evidence.

\paragraph{Synthesis.}
The audited literature yields three cross-benchmark takeaways. Position-aware evaluation is comparatively mature for retrieval-like tasks: NIAH and RULER make position explicit, and several later synthetic benchmarks continue that line. Mainstream reasoning and long-document understanding benchmarks, by contrast, remain much less explicit about position: SCROLLS, ZeroSCROLLS, L-Eval, LongBench, LooGLE, and 100-LongBench all evaluate long-context behavior on realistic inputs without systematically reporting where the relevant information appears. The existing synthetic reasoning exceptions do not close this gap: BABILong includes supplementary positional analysis and LongReason includes a binary inquiry-placement ablation, but neither provides a cross-benchmark audit or a joint treatment of position, filler content, and context length. Prior work such as Attribute or Abstain \citep{buchmann2024attribute}, LongPiBench \citep{tian2025longpibench}, and NoLiMa \citep{modarressi2025nolima} reinforces the broader point that position can be studied directly, but in retrieval or attribution settings rather than as a systematic audit of mainstream reasoning benchmarks. Pos2Distill \citep{wang2025pos2distill} is relevant at the conceptual level because it distinguishes retrieval-side and reasoning-side manifestations of position effects, but it does so in the context of mitigation rather than benchmark design; CRE is complementary in that it does not claim the conceptual split but tests whether benchmark design operationalizes that split in practice.

\section{Vendor Model Card Audit}
\label{app:vendor-audit}

This appendix reports the benchmark disclosure audit referenced in \cref{sec:intro}. For each of the four flagship long-context releases audited (DeepSeek-V4-Pro, MiMo-V2.5-Pro, Kimi-K2.6, GLM-5.1), we record whether nineteen benchmarks across three categories appear in the official vendor disclosure surface (tech report, model card, release blog, or platform documentation).

\subsection{Audit Method and Sources}
\label{app:vendor-audit-method}

The audit was performed on 2026-05-10 against vendor-official sources. The audit reflects vendor disclosures as of this date.
\begin{itemize}
\itemsep0em
\item \textbf{DeepSeek-V4-Pro}: tech report \url{https://huggingface.co/deepseek-ai/DeepSeek-V4-Pro/blob/main/DeepSeek_V4.pdf}, model card \url{https://huggingface.co/deepseek-ai/DeepSeek-V4-Pro}, release announcement \url{https://api-docs.deepseek.com/news/news260424}.
\item \textbf{MiMo-V2.5-Pro}: HuggingFace model card \url{https://huggingface.co/XiaomiMiMo/MiMo-V2.5-Pro}, official landing page \url{https://mimo.xiaomi.com/mimo-v2-5-pro}.
\item \textbf{Kimi-K2.6}: tech blog \url{https://www.kimi.com/blog/kimi-k2-6}, HuggingFace model card \url{https://huggingface.co/moonshotai/Kimi-K2.6}, platform documentation \url{https://platform.kimi.ai/docs/guide/kimi-k2-6-quickstart}.
\item \textbf{GLM-5.1}: HuggingFace model card \url{https://huggingface.co/zai-org/GLM-5.1}, GLM-5 base tech report \url{https://arxiv.org/abs/2602.15763}, Z.ai dev documentation \url{https://docs.z.ai/}.
\end{itemize}

Each benchmark-vendor cell receives one of three verdicts:
\begin{itemize}
\itemsep0em
\item \textbf{M (Main)}: appears prominently in the deployed product's main result or headline comparison table with a metric value.
\item \textbf{A (Ablation)}: appears only in a base-model evaluation, architecture ablation, or proxy-model ablation (e.g., 9B-proxy for design choice justification), not in the deployed product's main result-table.
\item \textbf{---}: absent from all audited official sources.
\end{itemize}

\subsection{Benchmark Coverage Matrix}
\label{app:vendor-audit-matrix}

\begin{table}[H]
\centering
\footnotesize
\setlength{\tabcolsep}{3pt}
\begin{tabular}{l l c c c c}
\toprule
\textbf{Benchmark} & \textbf{Group} & \textbf{DeepSeek-V4-Pro} & \textbf{MiMo-V2.5-Pro} & \textbf{Kimi-K2.6} & \textbf{GLM-5.1} \\
\midrule
NIAH                       & A & --- & --- & --- & A (9B-proxy) \\
RULER                      & A & --- & --- & --- & A (9B-proxy) \\
LongBench / V2             & A & A V2 51.5 (Base) & --- & --- & --- \\
HELMET                     & A & --- & --- & --- & A (9B-proxy) \\
$\infty$Bench              & A & --- & --- & --- & --- \\
BABILong                   & A & --- & --- & --- & --- \\
LOFT                       & A & --- & --- & --- & --- \\
\midrule
SWE-Bench Verified         & B & M 80.6 & M 78.9 & M 80.2 & --- \\
SWE-Bench Pro              & B & M 55.4 & M 57.2 & M 58.6 & M 58.4 \\
ClawEval                   & B & --- & M 64.0 & M 80.9 & --- \\
GDPVal                     & B & M 1554 Elo & M 1581 Elo & --- & A (AA Index v4.0) \\
BrowseComp                 & B & M 83.4 & --- & M 83.2 & M 68.0 \\
Terminal-Bench 2.0         & B & M 67.9 & M 68.4 & M 66.7 & M 63.5 \\
TAU-bench / $\tau^3$       & B & --- & M $\tau^3$ 72.9 & --- & M $\tau^3$ 70.6 \\
\midrule
AIME (2024/2025)           & C & --- (HMMT 95.2) & A Base 37.3 & --- (2026: 96.4) & --- (2026: 95.3) \\
GPQA-Diamond               & C & M 90.1 & M 66.7 & M 90.5 & M 86.2 \\
MMLU / MMLU-Pro            & C & M Pro 87.5 & M Pro 68.5 & --- & A Base 61.2 \\
LiveCodeBench              & C & M 93.5 & A Base 39.6 (v6) & M v6 89.6 & --- \\
HLE                        & C & M 37.7 & M 13.3 & M 54.0 & M 31.0 \\
\bottomrule
\end{tabular}
\caption{Vendor model card audit covering nineteen benchmarks across four flagship long-context releases. \textbf{M} = disclosed in the deployed product's main result-table; \textbf{A} = disclosed only in an ablation, base-model, or proxy-model table; \textbf{---} = absent from audited sources. Metric values shown where available. Group A: long-context positional evaluation. Group B: agent / coding. Group C: general capability.}
\label{tab:vendor-audit}
\end{table}

\subsection{Coverage Summary}

\paragraph{Group A (long-context positional, 7 benchmarks $\times$ 4 vendors = 28 cells).}
Zero of the 28 cells appear in deployed-product main result-tables. Four cells appear in ablation or proxy-model tables: DeepSeek-V4-Pro reports LongBench-V2 (51.5 EM) only in the Base eval table; GLM-5.1 reports NIAH, RULER, and HELMET inside a 9B-proxy architecture ablation justifying sparse attention design choices. The remaining 24 cells are entirely absent from official vendor disclosure.

\paragraph{Group B (agent / coding, 7 benchmarks $\times$ 4 vendors = 28 cells).}
20 of the 28 cells appear in main result-tables (counting $\tau^3$ entries as TAU-bench family). SWE-Bench Pro and Terminal-Bench 2.0 are disclosed by all four vendors. SWE-Bench Verified is disclosed by three of the four (GLM-5.1 substitutes Pro). BrowseComp is disclosed by three of the four (MiMo-V2.5-Pro absent). ClawEval and GDPVal each disclosed by two vendors. The original TAU-bench is absent from all four; MiMo-V2.5-Pro and GLM-5.1 report $\tau^3$ as a TAU-bench family substitute.

\paragraph{Group C (general capability, 5 benchmarks $\times$ 4 vendors = 20 cells).}
GPQA-Diamond and HLE are disclosed by all four vendors in main result-tables. MMLU/MMLU-Pro is disclosed by three (one in ablation only). LiveCodeBench is disclosed by three (one in ablation only). Canonical AIME (2024 or 2025) is absent from all four vendor main result-tables; substitutions include HMMT 2026 (DeepSeek-V4-Pro), AIME-2026 (Kimi-K2.6, GLM-5.1), and AIME-2024/25 reported only on the base model (MiMo-V2.5-Pro).

\paragraph{Aggregate.}
The disclosure asymmetry described in \cref{sec:intro} is reflected in the aggregate counts: long-context positional coverage at 0/28 main result-table cells, agentic/coding coverage at 20/28 main result-table cells. These audit data ground the vendor disclosure claim in \cref{sec:intro}. They complement the positional-sensitivity findings in \cref{sec:results} (initial five-model set) and \cref{sec:discussion-round2} (four newer releases).

\end{document}